\documentclass{article}

\usepackage{arxiv}

\usepackage{graphicx}

\usepackage{moreverb,url}
\usepackage[colorlinks,bookmarksopen,bookmarksnumbered,citecolor=red,urlcolor=red]{hyperref}

\begin{document}


\title{Graph Input Representations for Machine Learning Applications in Urban Network Analysis}

\author{
Alessio Pagani \\ The Alan Turing Institute, UK \\ apagani@turing.ac.uk  \And
Abhinav Mehrotra \\ University College London, UK \\ a.mehrotra@ucl.ac.uk \And
Mirco Musolesi \\ University College London, UK \\The Alan Turing Institute, UK \\ University of Bologna, Italy \\ m.musolesi@ucl.ac.uk
}
%

\date{}

\maketitle

\begin{abstract}
Understanding and learning the characteristics of network paths has been of particular interest for decades and has led to several successful applications. Such analysis becomes challenging for urban networks as their size and complexity are significantly higher compared to other networks. The state-of-the-art machine learning (ML) techniques allow us to detect hidden patterns and, thus, infer the features associated with them. However, very little is known about the impact on the performance of such predictive models by the use of different input representations.

In this paper, we design and evaluate six different graph input representations (i.e., representations of the network paths), by considering the network's topological and temporal characteristics, for being used as inputs for machine learning models to learn the behavior of urban networks paths.
%
The representations are validated and then tested with a real-world taxi journeys dataset predicting the tips using a road network of New York. Our results demonstrate that the input representations that use temporal information help the model to achieve the highest accuracy (RMSE of 1.42\$).
\end{abstract}

\keywords{Urban Networks, Graph Learning, Path Representation}

\section{Introduction}

Numerous important problems can be studied using the conceptual and theoretical framework of network science. Several structure and topological properties of networks have been widely studied in the recent years~(\cite{10.1109/CICSyN.2011.40,Gross:2005:GTA:1212553,Bondy:1976:GTA:1097029,graph_theory:395714}).
One of the most basic concepts in network science is the definition of network path~(\cite{7538314,Ahuja:2017:NFT:3153716}), i.e., a sequence of edges that joins a sequence of edges. In the case of a finite path, it is possible to identify the origin and the destination of a path as the starting and ending node of the sequence. This concept is widely used to study public infrastructures and utilities, such as trajectories and traffic flows (\cite{1617378,VanWoensel:trafficflow}), social networks (\cite{Srinivasan:socialnet,Kwak:2010:TSN:1772690.1772751}), ecological networks (\cite{dale_2017}), just to name a few. More recently, researchers have been focusing on the problem of classifying paths (\cite{RAMIREZCORONA2016179,Papa:2009:SPC:1541827.1541837,paths2:Charu,Jiang:2016:CSB:2936311.2746403}).
For example, in the field of urban networks, classifying the paths (e.g., detecting the mean of transport) and predicting related characteristics (e.g., cost of a journey) could be useful from different prospectives. Fields of interest concern travel planning or environmental and social analysis, such as classification of urban areas, for example for understanding their wealth and wellbeing level. 

%

Recently, machine learning (ML) techniques~(\cite{ml1,ml2,ml3}) have been applied to the analysis of urban networks~(\cite{Duvenaud:molecularFingerprints,structuralRNN,Lee:2009:KBR:1497653.1498379}). However, due to memory and computational limitations, performing analysis on the networks remains challenging as their dimension increases (such as in the case of the road or transport networks of large cities). This is also due to the fact that, unlike other types of data that are easily transformed for example into time series or grids (e.g., an image can be represented as a matrix of pixels), there is no standard way of representing the network paths for using them as input to ML models. When the training dataset is large in size, ML algorithms such as Random Decision Forests or Deep Neural Networks, require a significative amount of memory and computational power to train their models (\cite{Kearns:1990:CCM:76476,Jordan255}). For this reason, the traditional path representations (such as adjacency matrix) could be used only when the networks are small (their size increase exponentially with the size of the networks). 

In this paper, we investigate the impact of different \textit{network path representations} that can be exploited as inputs to effectively train ML models.
More specifically, we designed and evaluated 6 different network path representations, by considering their topological and temporal characteristics, for performing ML-based classification and regression to predict characteristics associated to the paths, i.e., the value of a \textit{linked features} (e.g., travel time, cost of a ride).
As a proof of concept and for evaluating the proposed approach, in this paper we compute the tip of the taxi journeys using only network related features (excluding ride related information, such as cost of the ride, number of people).
%
In particular, we consider the problem of predicting the tip of a trip by considering a real-world case study, namely the analysis of the prediction of trips contained in the New York Taxi dataset (\cite{TaxiDataset}). More specifically, the prediction task that we consider is the estimation of the amount of the tip of each ride by using in input only characteristics of the road network. In particular we compare two popular algorithms for this task, i.e., \textit{Random Decision Forest} (\cite{Ho:1995:RDF:844379.844681}) and \textit{Deep Neural Networks} (\cite{DL1,SCHMIDHUBER201585}). 

The key contributions of this paper can be summarized as follows:
\begin{itemize}
\item We propose 6 methods, suitable for large-scale urban networks, to efficiently represent large-network paths to be used as inputs for machine learning algorithms. Each method is characterized by a different level of path compression and type of feature selection (e.g., preserving time and topological features).
\item We demonstrate that the input representations, used as input to train Random Decision Forests and Deep Neural Networks, are effective in classifying paths according to some shared properties (e.g., paths with similar cost or tip value).
We evaluate the effectiveness of these representations by considering a real-world taxi trip data set.
\end{itemize}

Our results show that the ML model's performance exploiting different input representations vary according to the characteristics of the paths, such as the part of the network (i.e., neighborhoods) they cross or the importance of the origin and destination points (e.g., a node close to an airport is generally more important). 
For example, when the path characteristic we want to predict is simple (e.g., the distance of two points in a path), even a basic representation (e.g., a list of nodes) performs well. On the other hand, in more complex scenarios, considering the specific information about origin and destination nodes and the network topology (e.g., neighbor nodes) is fundamental to achieve high prediction accuracy. Moreover, the results of the experiments with the New York Taxi dataset demonstrate that the value of tips could be predicted with high accuracy (RMSE 1.42\$). We also observe that, for this analysis, the input representations that exploit the network topology are more effective compared to the representations that use the temporal information.

\section{Designing Graph Input Representations}

\begin{figure}[!htb]
     \centering
     \includegraphics[width=\textwidth]{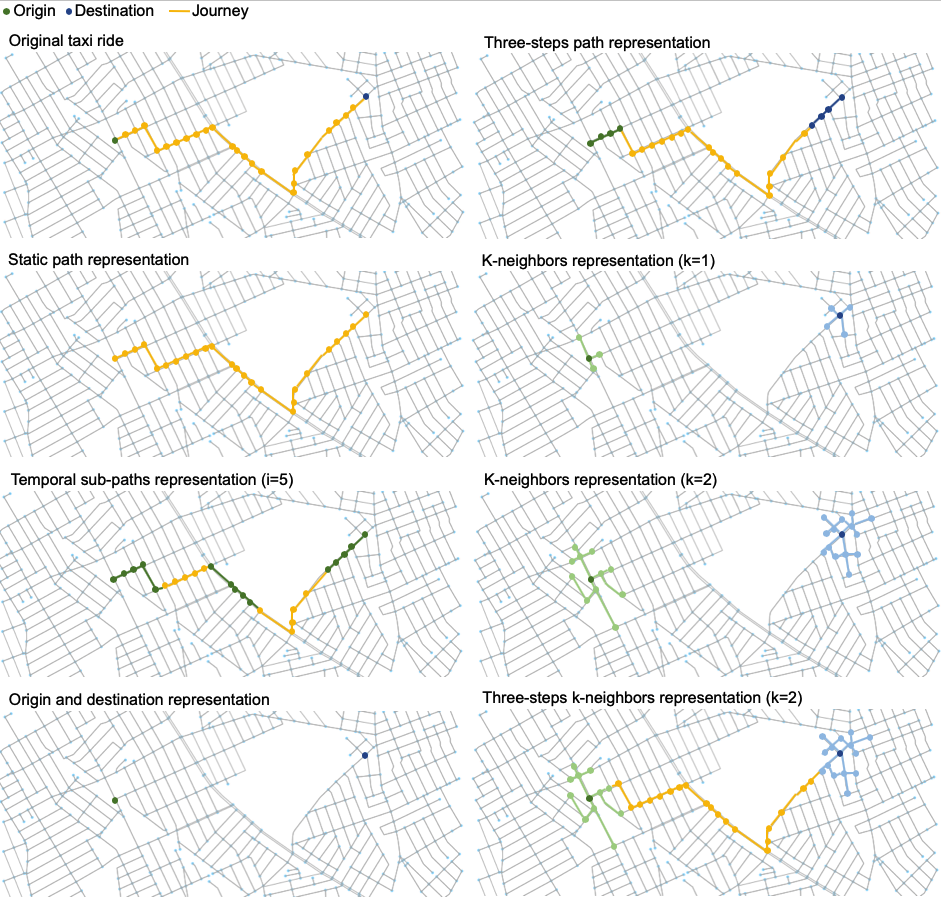}
     \caption{Urban network path representations for ML, each representation has different properties and advantages.}
     \label{fig:ML-representations}
\end{figure}

\subsection{Representation of a Path}
In this section we discuss the design of the path representations $I$ (i.e., journeys in the city, public transport lines, traffic flows) that, associated to a variable $y$ (e.g., price, time), can be used for training ML models. The ML models can then be used to predict the output variable $y$ of new paths.

All the proposed input designs rely on a basic representation of the nodes' position: the \textit{node vectors}. A \textbf{node vector} in a network of $N$ nodes is defined as an ordered list of size $N$, where each position in the vector corresponds to a specific node in the graph. 
To represent a path in a node vector, we set elements of this vector corresponding to nodes in the path as 1 (i.e., activated nodes), the rest of the vector elements are set as 0 (i.e., not activated nodes). 
Consequently, for any input representation of a network path, a node vector will have at least two nodes set to $1$ (i.e., origin and destination nodes in the case of a path of length 2). In urban networks, paths can represent journeys in the city, public transport lines, traffic flows and so on.

The simplest representation of a network path is with only one \textit{node vector} containing all the nodes of the path. Overall, we focus on designing graph input representations that are capable of compressing the size of paths while preserving the temporal, geographic and semantics information. 
In order to further reduce the dimensionality of the input we design representations that take into consideration only the most informative parts of a path (e.g., in paths where origin and destination points are semantically relevant places, the rest of the path may be discarded without information loss) and the topology of the network.

As shown in Figure \ref{fig:ML-representations}, different representations can compress the paths in different ways, preserving different types of relevant information. For example, the \textit{origin and destination} representation uses only origin and destination points,  highlighting possible relationships between the paths and the semantics of the places; the \textit{static path} representation compresses all the information in a single vector, but it does not contain any other specific information; the \textit{temporal sub-paths} representation divides the static path in an ordered sequence of sub-paths;
the \textit{three-steps path} representation is focused on the semantic of the places and on the temporal sequence of the nodes; the \textit{k-neighbors} representations (1-neighbor and 2-neighbors in Figure \ref{fig:ML-representations}) consider mainly the geographic information, extending the representation to the surrounding area of the origin and destination of the paths. Finally, the \textit{three-steps k-neighbors} representation merges the three-steps path and the k-neighbors representations.

To improve the information available, combination of those representations are also analyzed in the following sections.

\subsection{Static Path Representation}
In the static path representation, the path representation consists of a single node vector in which each element corresponds to a node in the network. The element is set to 1 if the node belongs to the path, 0 otherwise (Figure \ref{fig:ML-representations} (c)). More formally, given a graph $G = [V, E]$ with $V$ as:

\begin{equation}  \label{eq:general_network}
V = [v_0, \dots, v_{N-1}]
\end{equation}

\noindent
and a path \textbf{P} of length $T$ (i.e., a sequence of $T$ vertices) defined as:

\begin{equation}  \label{eq:general_path}
 \textbf{P} = [v'_0, \dots, v'_j,\dots, v'_{T-1}]
\end{equation}

\noindent
then, the resulting representation is the node vector $I_{static}$, which is defined as:
\begin{equation} \label{eq:non_temporal}
\textbf{I}_{static}=[i_{v_0}, \dots, i_{v_{N-1}}]
\end{equation}

\noindent
where:

$\forall j \in [0, N)$,  $i_{v_j} = 1$ if $v_j \in  \textbf{P}$, $i_{v_j} = 0$ otherwise. \\

\noindent
For example, given a network $G$ with 5 nodes and a path $\textbf{P}=[v_2,v_3,v_5]$ composed of 3 nodes, the \textit{static path} representation is:

$ \textbf{I}_{static} = [0,1,1,0,1]$.

\subsection{Temporal Sub-paths Representation}
Temporality, the order of the nodes in the paths, is often an important path feature. In predictive tasks where this is important, we propose this technique to include temporality.
The \textit{temporal sub-paths} representation embeds the temporal information of network paths by dividing each path into temporally ordered sub-paths.
A sub-path is a subsequence of the original path, created by splitting it in ordered sub-sequences (i.e., the first $i$ nodes in the path are assigned to the first sub-path, the second $i$ nodes to the second one and so on).
Since a sub-path contains a subsequence of the original path, using the same logic of the \textit{static path} representation, we convert each sub-path in a node vector by setting to 1 only the path nodes that belong to that sub-path. The final result is a list of node vectors, each one representing a part of the original path. 

The lower is the number of nodes in each sub-path, the higher is the number of sub-paths required to represent a full path and thus the accuracy of the temporal information: in the extreme case of sub-paths with one element, all the temporal information is preserved. 
However, the higher is the number of sub-paths and the bigger is the size of the representation, consequentially increasing the training cost of the ML models (in terms of computational resources and time).

More formally, given a path $\textbf{P}$ (see Eq. \ref{eq:general_path}), this input representation is the concatenation of $S$ sub-paths ($\textbf{p}_{s}$), where $s \in [0, \dots , S-1]$ and a sub-path $\textbf{p}_{s}$ is an ordered sub-sequence of the original path $\textbf{P}$ containing $N_S$ nodes. 
More specifically, a sub-path is defined as:
\begin{equation}  \label{eq:sub-path_path}
 \textbf{p}_{s} = [v'_{N_S*s}, v'_{N_S*s+1}, \dots, v'_{N_S*s+(N_S-1)}]
\end{equation}

\noindent
where:

$N_S$ is the number of path nodes in each sub-path.\\

\noindent
Finally, each sub-path ($\textbf{p}_{s}$) is converted to a node vector ($\textbf{I}_{s}$) as:
\begin{equation}  \label{eq:sub-path_nodevector}
 \textbf{I}_{s} = [i_{v_0}, \dots, i_{v_{N-1}}]
\end{equation}

\noindent
where:

$\forall j \in [0, N)$,  $i_{v_j} = 1$ if $v_j \in  \textbf{p}_{s}$, $i_{v_j} = 0$ otherwise. \\

It is worth noting that if $N_S$ is not a divisor of $T$, the last sub-path would not contain exactly $N_S$ nodes, instead it would comprise of the remainder ones (i.e., $T \% N_S$ nodes).

\noindent
Since one of our aims is to limit the size of the input, we set a limit on the maximum number of sub-paths. For simplicity, if a path exceeds the maximum number of sub-paths, the central ones are removed. The ratio behind this choice is that we hypothesize that the parts of the journeys with less information are the ones in the center, as opposed to the origin and destination parts that contain more semantic and geographic information.

Moreover, as also discussed earlier, we assume that the origin and destination nodes may be more relevant than other nodes, and, therefore, indicating them might be helpful for the network. For this reason the first and last node vectors in the final input representation contains origin and destination nodes respectively, and then followed by all the other node vectors in temporal order. 

The resulting input representation is the following:
\begin{equation} \label{eq:temp_sub-paths}
 \textbf{I}_{TC}=\textbf{I}_{0} \sqcup  \textbf{I}_{S-1} \sqcup  \textbf{I}_{1} \sqcup \dots \sqcup  \textbf{I}_{S-2}
\end{equation}
where:

$S$ is the maximum number of sub-paths.

\noindent
This input representation has always be of size $S * N$ (i.e., the number of sub-paths multiplied with the total element in a node vector). Therefore, if a shorter path can only be grouped into less sub-paths than $S$, all the following node vectors are all set to 0. On the other hand, if a path has more sub-paths the central ones are removed as we need to preserve the information about origin and destination. 

For example, given a network $G$ with 5 nodes and a path $\textbf{P}=[v_2,v_3,v_5]$ composed of 3 nodes, the \textit{temporal sub-path} representation with $S=3$ and $N_S=1$ is:

$\textbf{I}_0 = [0,1,0,0,0]$.

$\textbf{I}_1 = [0,0,1,0,0]$.

$\textbf{I}_2 = [0,0,0,0,1]$.

$\textbf{I}_{TC} =  \textbf{I}_{0} \sqcup \textbf{I}_{2} \sqcup \textbf{I}_{1} =  [0,1,0,0,0,0,0,0,0,1,0,0,1,0,0]$.

\subsection{Origin and Destination Representation}
This input representation is a variant of the \textit{static path} representation used when the full path is not available (e.g., enter and exit points of passengers in the underground network) and the shortest path (or any other technique to reconstruct the path) does not accurately reflect the condition of the actual path.

This representation is composed of two node vectors, one for the origin and one for the destination node of a network path (Figure - \ref{fig:ML-representations} (g)).
Compared to the \textit{temporal sub-paths} representation, this representation has a significantly smaller size (it is always $2*N$, one node vector for the origin and one node vector for the destination) and it can thus be used to represent in a compact manner also those paths whose origin and destination points are considerably more important than the others (semantic information).
Given a path $P$ (see Eq. \ref{eq:general_path}), the resulting input representation is:
\begin{equation} \label{eq:orig_and_dest}
\textbf{I}_{OD}=\textbf{I}_{start} \sqcup \textbf{I}_{end}
\end{equation}
where:

$\textbf{I}_{start}$ is the node vector that includes only the node $v_0$ (all nodes set to 0 except the origin node, set to 1). 

$\textbf{I}_{end}$ is the node vector that includes only the node $v_{T-1}$ (all nodes set to 0 except the destination node, set to 1).\\

For example, given a network $G$ with 5 nodes and a path $\textbf{P}=[v_2,v_3,v_5]$ composed of 3 nodes, the \textit{origin and destination} representation is:

$\textbf{I}_{start} = [0,1,0,0,0]$.

$\textbf{I}_{end} = [0,0,0,0,1]$.

$\textbf{I}_{OD}=\textbf{I}_{start} \sqcup \textbf{I}_{end} = [0,1,0,0,0,0,0,0,0,1]$.

\subsection{Three-Steps Path Representation}
This technique is an extension of the \textit{origin and destination} representation that, like that one, discriminates origin and destination of the journeys (temporal information) but, in addition, it includes the full path sequence of nodes (Figure - \ref{fig:ML-representations} (b)).
It is generally less accurate than the \textit{temporal sub-paths} representation, but it has a significantly smaller size and it is thus suitable for huge networks (big cities or big areas).

This representation is the concatenation of the \textit{origin and destination} representation and the \textit{non-temporal} representation.
The resulting representation has size $3*N$ and it is defined as follow:
\begin{equation} \label{eq:3-steps}
\textbf{I}_{3steps}=\textbf{I}_{OD} \sqcup \textbf{I}_{static}
\end{equation}

In the first node vector there is only the origin node set to 1 while in the second node vector only the destination node set to 1. Finally, in the third node vector, all the elements of the path are set to 1.\\

For example, given a network $G$ with 5 nodes and a path $\textbf{P}=[v_2,v_3,v_5]$ composed of 3 nodes, the \textit{three-steps path} representation is:

$\textbf{I}_{3steps}=\textbf{I}_{OD} \sqcup \textbf{I}_{static} = [0,1,0,0,0,0,0,0,0,1,0,1,1,0,1]$.

\subsection{K-Neighbors Representation} 
For some classification and regression problems it is important to exploit the network topology and to consider areas of the network instead of single nodes (e.g., in a road network, two points on the same road are generally more similar than two points at different boundaries of the city, even if their structural properties are similar).
The \textit{k-Neighbors} representation is an enriched version of the \textit{origin and destination} representation, with focus on the neighborhood of the the origin and destination (geographic information) (Figure \ref{fig:ML-representations} - (d) and (f)): given a node vector with one node activated (hereafter called 'base' node), the \textit{k-neighbors nodes} of this base node are also activated (set to 1) in the same vector. The \textit{k-neighbors nodes} of a node are its neighbors (nodes linked to it) until grade $k$.
For example, with $k = 1$ only the nodes directly connected to the 'base' node are included, with $k = 2$ also the neighbors of the neighbors of the base node are taken into account, and so on.
As one of the goals of this work is to handle large urban networks and the geographic information is related to the semantic information, in this input representation the only 'base' nodes selected are the origin and destination nodes (the final representation has size $2*N$).

Formally, given a path $\textbf{P}$ (see Eq. \ref{eq:general_path}), the k-neighbors representation is:
\begin{equation} \label{eq:k-neighbors}
\textbf{I}_{KN}=\textbf{I}_{k-start} \sqcup \textbf{I}_{k-end}
\end{equation}
where:

$\textbf{I}_{k-start}$ is the node vector that includes $v_0$ and its k-neighbors.

$\textbf{I}_{k-end}$ is the node vector that includes $v_{T-1}$ and its k-neighbors.\\

For example, given a network $G$ with 5 nodes and a path $\textbf{P}=[v_2,v_3,v_5]$ composed of 3 nodes, node $v_2$ connected to nodes $v_1$, $v_3$, $v_4$ and node $v_5$ connected only to node $v_4$.
The \textit{k-neighbors} representation is:

$\textbf{I}_{k-start} =  [1,1,1,1,0]$.

$\textbf{I}_{k-end} =  [0,0,0,1,1]$.

$\textbf{I}_{KN}=\textbf{I}_{k-start} \sqcup \textbf{I}_{k-end} = [1,1,1,1,0,0,0,0,1,1]$.

\subsection{Three-Steps K-Neighbors Representation}
Like the \textit{three-steps path} representation, this representation is a combination of the \textit{k-neighbors} representation and the \textit{non-temporal} representation. The aim of this representation is to consider the topology of the network (using k-neighbor), the temporal information (discriminating origin and destination) and the full path, without a significant memory consumption (the input representation has size $3*N$).
The resulting representation is:
\begin{equation} \label{eq:3-steps_k-neighbors}
\textbf{I}_{3steps-KN}=\textbf{I}_{KN} \sqcup \textbf{I}_{static}
\end{equation}

In this representation the first node vector indicates the origin node and its k-neighbors by setting them to 1. While the second node vector indicates the destination node and its k-neighbors by setting them to 1. Finally, the third node vector indicates all the nodes of the path by setting them to 1. \\ 

For example, given a network $G$ with 5 nodes and a path $\textbf{P}=[v_2,v_3,v_5]$ composed of 3 nodes, node $v_2$ connected to nodes $v_1$, $v_3$, $v_4$ and node $v_5$ connected only to node $v_4$.
The \textit{three-steps k-neighbors} representation is:

$\textbf{I}_{3steps-KN} = \textbf{I}_{KN} \sqcup \textbf{I}_{static} = [1,1,1,1,0,0,0,0,1,1,0,1,1,0,1]$.\\

\section{Evaluation Setup}

In this section we discuss our approach to evaluate the proposed input representations. We start with discussing the dataset used, then the construction of prediction models and, finally, we introduce the baselines and metrics for quantifying the performance of prediction models.
 
\subsection{NYC Taxi Dataset} 
\label{subsec:dataset}




People use taxis to travel from one part of a city to another. Along with the bill there is a tip offered to the drivers that reflects the willingness of users to pay some extra money for their trips.
Although the tip depend on several factors, we hypothesize that the network topology is key factor for determining it. Therefore,  we use a dataset of taxi trips for constructing a model that can predict the tip for each trip. The key motivation for using this dataset as a case study is that it can be used to improve the pricing policy of the private companies, i.e. by detecting the areas of the city with the highest willingness to pay at different times, setting the right fares or moving the cars in more profitable areas.

We use the \textit{yellow and green taxi trip dataset}~(\cite{TaxiDataset}) containing the information about 1 million taxi trips during a period of 12 months. The trip information included in the dataset is as follows: VendorID, pick-up date and time, drop-off date and time, passenger count, trip distance, pick-up longitude, pick-up latitude, RatecodeID, store and fwd flag, drop-off longitude, drop-off latitude, payment type, fare amount, extra, mta tax, tip amount, tolls amount, improvement surcharge, total amount. However, we consider only the pick-up and drop-off locations of trips to model our problem.
In this work we used the dataset of the first 6 months of 2016, using 1 million of rides in the area of Manhattan and Brooklyn. We selected these two areas in order to have a variety of business and residential areas, shopping and touristic areas as well as an airport. 
The journeys used for training and validating the models are filtered so as to obtain datasets with a uniform distribution of the tips. This assure that the models are general and able to predict all possible tip values, avoiding overfitting around the most common tip values.

In this work the urban network is built using the New York City road data provided by OpenStreetMap (OSM) (\cite{OSM-newyork}), that is composed of 20,990 nodes.
In order to create a network path for each trip, the pick-up and drop-off locations of that trip are associated to the closest node in the road network and the route of the trip is then estimated by computing the shortest path with the Dijkstra's algorithm~(\cite{Dijkstra:1959:NTP:2722880.2722945,Mehlhorn:2008:ADS:1404505}).

\subsection{Constructing Predictive Models}
\label{sec:construct_models}
We construct models for predicting tips with each input representation by using two machine learning algorithms: (i) Random Decision Forest (\cite{Ho:1995:RDF:844379.844681}) and (ii) Deep Neural Network (\cite{DL1,SCHMIDHUBER201585}). The rationale behind using these two algorithms is to test the proposed input representations with machine learning algorithms with different learning techniques and peculiarities.
We fine tune both models by optimizing their hyper-parameters as discussed below. 

\subsubsection{Tuning the Random Decision Forest Algorithm} 
\label{sec:tune_rf}
We tune the Random Decision Forest algorithm by optimizing the following three hyper-parameters: (i) number of trees, (ii) max depth of the trees, and (iii) minimum number of samples required to split an internal node or to be a leaf node. This is performed through a standard tuning function of the \textit{Scikit-learn} Python library (\cite{scikit-learn}).

\subsubsection{Tuning the Deep Neural Network Algorithm}
\label{sec:tune_nn}
In order to tune the deep neural network models there are two key hyper-parameters to be optimized: (i) the number of hidden layers, and (ii) the number of nodes in the hidden layers. However, the process of finding the optimal values for these parameters is an intricate task, consequently, we optimized the value by exploring the number of hidden layers between 1 and 5. Moreover, the number of nodes are set as follows: 

\begin{equation}\label{eq:nodes}
n_i = IL_{size} / (NDR * i) 
\end{equation}

Here, $n_i$ indicates the number of nodes in $i^{th}$ layer, $IL_{size}$ refers to the size of input layer, and $NDR$ refers to \textit{node division ratio}, which is optimized $\in [4,8,16,32,64]$. 

Another parameter we consider for optimization is the \textit{dropout ratio} (\cite{JMLR:v15:srivastava14a}), which is used to prevent overfitting the models. During training, random nodes are dropped, along with their connections, from the neural network generating different \textit{thinned} networks. During test phase, an \textit{unthinned} neural network (without dropout) is used. We performed the evaluation by using values of dropout ratio between 0\% to 50\% with a step of 10\%.

\subsection{Evaluation Criteria}

In order to quantify the performance of models, we compute the \textit{root mean-squared error} (RMSE) for the test predictions. 
Moreover, we use the \textit{k-fold cross validation} technique for validating the models because it significantly reduces bias as most of the data is used for fitting, and also significantly reduces variance as most of the data is also being used in validation set.
The folds are constructed by dividing the dataset in $k$ folds, using $k-1$ folds to train the models and the remaining one for validation. The experiments are repeated $k$ times, in order to use each possible combination of training and validation folds, and the average of the outcomes is used as final result.

In order to examine a model's performance we compare it with two baselines: \textit{overall average tip} and \textit{area average tip}. 
The \textit{overall average tip} baseline is the simplest case in which the average of the overall tips in the dataset is used as a prediction output. The \textit{area average tip} baseline takes the average of tips for all rides with similar departure and destination areas and uses it as prediction output. Here, an area is defined as a circle with radius $r$ and center in the origin or destination of the considered ride.
For the taxi dataset we selected the optimal radius $r$ by testing different values from 0.1 km to 5 km (optimal $r$ is 1 km).


\section{A Case Study: Predicting the Tip for Taxi Rides}
\label{sec:results}

In this section we present the results for the tip predicting task for given taxi trips, using the \textit{NY Taxi Journeys} dataset\footnote{We firstly performed an extensive analysis to measure the robustness of the proposed input representations using synthetic networks. Our results show that all the representations lead to accurate predictions even in presence of noise. The analysis can be found in the Supplementary Material.}. 
In a taxi ride, the quantity of tip left depends on several factors, the main ones are the travel time and distance. In this case study we argue that there are other latent factors that influence the amount of tips. In order to investigate this, we try to predict the tips left in New York City (Manhattan and Brooklyn) using exclusively the urban network graph and representing the journey path (from the origin to the destination) through the input representations defined above. We did not include the geographic distance of a journey (i.e., the sum of the length of the edges between all traversed nodes) because this would bias the results by possibly training models that compute the total length of a path to predict the tip, which is already known to be a factor that determines it.
The objective is to extract information without calculating any secondary metrics calculated from the graph (e.g., distance between origin and destination). We use this approach to design general and scalable models. As a result, the input representations can be adopted in a wide range of applications where graph learning is required.

We examine the effectiveness of these input representations for training two types of ML algorithms: Random Forest and Deep Neural Network. We chose these two algorithms because they are among the most widely used for this kind of regression problems. Moreover their comparison represents an interesting case of study: both models tend to be very powerful - yet surprisingly uncorrelated. As such, analyzing the two approaches proves the general effectiveness of the proposed input representations. Finally, both algorithms are used to construct models by using each input representation as the input, which enables us to compare the effectiveness of individual representations for enabling a model to predicting the tip for journeys. Moreover, it is worth noting that we fine-tuned models for both ML algorithms by using the grid search approach. 
Furthermore, we use the $k$-fold cross validation approach (with $k=20$) to statistically validate of our results.

\begin{figure}[t]
      \centering
      \includegraphics[width=0.6\columnwidth]{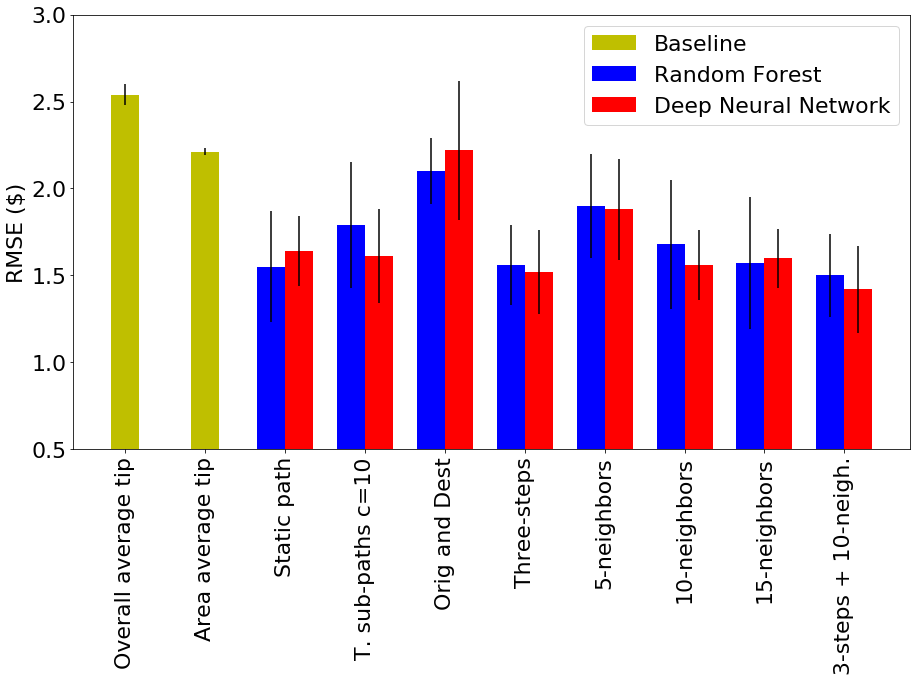}
      \caption{Tip prediction: comparison of the input representations, Random Forest and Deep Neural Network models are trained with all the representations and their performance are compared.}
       \label{fig:tip-comparison-results}
\end{figure}

In Figure \ref{fig:tip-comparison-results} we compare the performance of models trained using the different input representations with both Random Forest and Deep Neural Network. We use the \textit{overall average tip} (RMSE 2.54\$) and \textit{area average tip} (RMSE 2.21\$) as baselines for prediction performance. The results demonstrate that the full path itself (i.e., \textit{non-temporal} representation) achieves RMSE 1.55\$. However, adding temporality to the input representation (i.e., using (i) the \textit{temporal sub-paths} representation and (ii) the \textit{3-steps} representation) the performance of models do not improve (i.e., RMSE of 1.61\$ and 1.52\$ respectively). This could be due to the fact that people going from place A to place B leave a similar tip as people going from place B to place A.

On the other hand, when we use only origin and destination nodes of a path (i.e., \textit{origin and destination} representation) the trained models do not perform better than the baselines (RMSE 2.10\$). This depicts that this representation does not contain enough information for effectively learning the path behavior of huge graphs. However, we observe that by considering the network topology along with origin and destination nodes (i.e., using the \textit{k-neighbours} representation) the performance of models can be improved compared to using only information about the origin and destination nodes. More specifically, our models using the \textit{k-neighbours} representation with $k$ as 10 and 15 achieve the RMSE of 1.66\$ and 1.57\$ respectively, which is comparable to the performance of models using the \textit{non-temporal} representation. 
Overall, we observe that the best performance (RMSE 1.42\$) is achieved by using \textit{3-steps k-neighbor} representation (i.e., when we concatenate the \textit{k-neighbor} representation and the \textit{non-temporal} representation) for constructing the models. 

For completeness, we compared our results also with models that use the distance feature, a parameter already known to be correlated to the value of a tip: these models reach an accuracy comparable to the models trained using our best input representation (RMSE approximately equal to 1.51\$). It is worth noting that the main difference is that our approach is not based on engineered features. 

It is worth noting that the observations regarding the performance of the input representations could be specific to this prediction task. We believe that all proposed input representations should be examined in other types of prediction tasks to validate their performance for those cases. In fact, depending on the task, the relevant information could be different and thus also the best input representation. For example, predicting the fare of trips might require more temporal information, while detecting areas where people go during leisure time might require topological information.

The impact of the DNN hyper-parameters configuration is discussed in Appendix 1 (see Supplementary Material).

\subsection{Trip Fare Prediction}
Tip prediction has been chosen as the main case study because this is a complex task that depends on many factors and it represents an exemplar application for the proposed input representations. As a comparison, in this section we discuss the input representations performance with another regression task. Using the same path representations, we trained different ML models to predict trip fares using only the road network and the trip paths. This task is usually accomplished using approaches computationally less expensive, as trip fares are mainly a combination of distance and time. However, we performed this analysis to prove that our approach is general and that it can be applied to different types of regression tasks, with the advantage that it can also be used for in-depth graph-based socio-economic analysis of cities.
%
%
%

\begin{figure}[t]
      \centering
      \includegraphics[width=0.6\columnwidth]{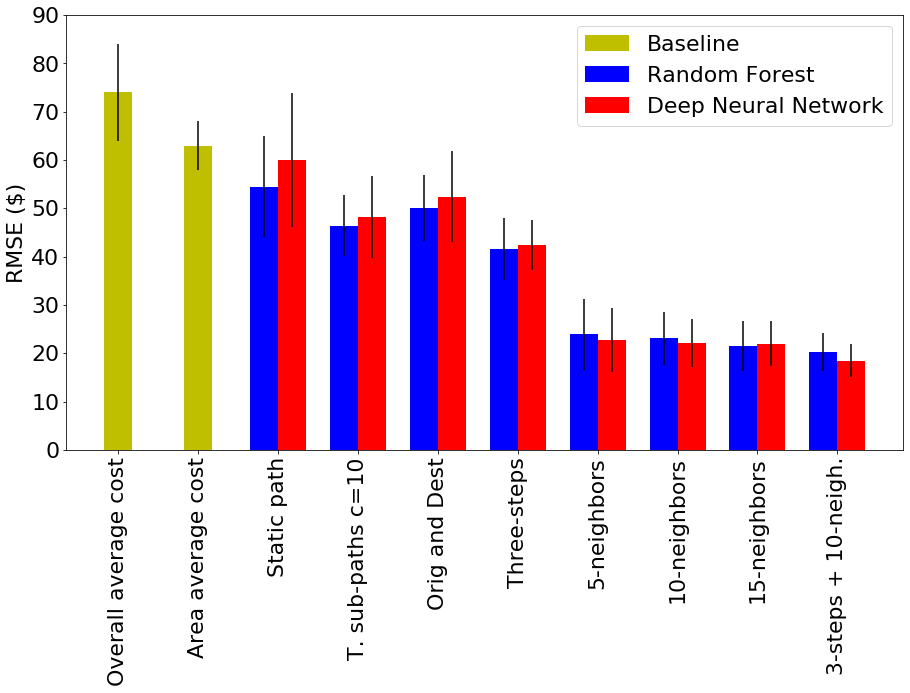}
      \caption{Fare prediction: comparison of the input representations, Random Forest and Deep Neural Network models are trained with all the representations and their performance are compared.}
      \label{fig:fare-comparison-results}
\end{figure}

Our results show that, similarly to the main case study, Neural Networks and Random Forests models have similar performance.
As expected, the length of the path influences the final cost of a journey and representations that include the full path perform better that the baselines (overall average cost RMSE 74\$, area average cost RMSE 63\$): static path, temporal sub-paths and three-steps representations obtain RMSE values between 40\$ and 50\$.
The models trained with k-neighbors representations (RMSE around 20\$), surprisingly, obtain a significant improvement compared to the full path ones. These results suggest that k-neighbors representations, having the ability to detect specific areas, identify specific places (e.g., places that can be reached only using toll roads, like airports).
The best results, as in the main case study, are obtained using the models trained with three-steps k-neighbors representations (RMSE 18.5\$).
A summary of the results is shown in Figure \ref{fig:fare-comparison-results}.

\subsection{Performance Analysis}

In this section we compare the performance of the different path representations in terms of time required to train a single model and memory used. We are interested in the most computationally-expensive model, i.e., Deep Neural Networks. The graph is composed of 20,990 nodes and the models are trained on a machine with the following setup: Ubuntu 16.04, Intel Xeon CPU E5-2690 v3 2.60GHz,  56GB RAM, GPU Tesla K80 24 GB GDDR5.

Our analysis (Table \ref{table:cost}) shows that  the fastest representation is the static path representation (that uses only one node vector): it requires around 40 minutes to train a Deep Neural Network model. Using two node vectors (origin and destination and k-neighbors representations) doubles the training time, while the three-steps path and three-steps k-neighbors representations (3 node vectors) require, on average, 135 minutes. The most computationally intensive representations are the temporal sub-paths that, depending on the number of sub-paths, use up to 850 minutes for training a single model. From a memory usage perspective, the resource requirement has a pattern similar to time: generally, a smaller number of vector nodes require less memory (2.4GB for the static path representation). However, it is worth noting that, thank to software optimizations, even when the number of node vectors is the same, sparser matrices require less memory. For example, considering the k-neighbors representation,  the memory required for the case of 5 neighbors is 4.1 GB. This raises to 5GB for 10 neighbors and it reaches 9GB when 15 neighbors are considered. 

\begin{table}[!ht]
\centering
 \begin{tabular}{|c c c|} 
 \hline
 Path representation & Time (min) & Memory (GB) \\  \hline
 Static path & 40 & 2.4 \\  \hline
 Temporal sub-paths (c=5) & 850 & 53  \\ \hline
 Temporal sub-paths (c=10) & 420 & 28  \\ \hline 
 Origin and destination & 80 & 2.4 \\ \hline
 Three-steps path & 135 & 3.7 \\ \hline
 K-neighbors (k=5) & 80 & 4.1 \\ \hline
 K-neighbors (k=10) & 80 & 5 \\ \hline
 K-neighbors (k=15) & 80 & 9.1 \\ \hline
Three-steps k-neighbors (k=5) & 135 & 5.7 \\ \hline
Three-steps k-neighbors (k=10) & 135 & 6.6 \\ \hline
Three-steps k-neighbors (k=15) & 135 & 10.8 \\ \hline
\end{tabular}
\caption{Performance comparison: time and memory required to train a single deep neural network. System used: Ubuntu 16.04, Intel Xeon CPU E5-2690 v3 2.60GHz,  56GB RAM, GPU Tesla K80 24 GB GDDR5.}
\label{table:cost}
\end{table}

\section{Related Work}


In this section we review the related work in two key areas, namely the studies about design of ML techniques based on graph inputs and some examples of ML approaches already used in urban network analysis.

\subsection{Graph Input Representations} 
Several graph embedding algorithms have been proposed in the past to generate vector representations for prediction tasks~(\cite{Compagnon:graph_embedding,Cox:10.2307/2983890,Perozzi:2014:DOL:2623330.2623732,Grover:2016:NSF:2939672.2939754}). Ideally, adequate embeddings should not suffer a loss of information compared to the original graph. One of the most commonly used techniques is to obtain the decomposition of a network through clustering algorithms. For instance, one-vs-rest logistic regression( \cite{Cox:10.2307/2983890}) (i.e, training $N$ different classifiers, each one designed for recognizing only a specific class) was used in relevant works such as \textit{DeepWalk}~(\cite{Perozzi:2014:DOL:2623330.2623732}) and \textit{Node2Vec}~(\cite{Grover:2016:NSF:2939672.2939754}). DeepWalk~(\cite{Perozzi:2014:DOL:2623330.2623732}) uses local information obtained from truncated random walks to learn latent representations by treating walks as the equivalent of sentences. In Node2Vec~\cite{Grover:2016:NSF:2939672.2939754} continuous feature representations are learnt for nodes in networks to map them to a low-dimensional space preserving nodes neighborhoods. 

Graphs inputs in neural networks have been studied by
~\cite{gori} and 
by~\cite{scarselli} as a form of recurrent neural networks. The authors use repeated application of contraction maps as propagation functions until node representations reach a stable fixed point. 

A propagation rule that uses convolution-like operations on graphs and methods for graph-level classification has been proposed by 
\cite{Duvenaud:molecularFingerprints}, which requires to learn node degree-specific weight matrices which does not scale to large graphs with wide node degree distributions. 
This limitation has been overcome by 
\cite{Kipf:semusupervisedClassifiation}, they propose a model that uses a single weight matrix per layer and deals with varying node degrees through an appropriate normalization of the
adjacency matrix.
In contrast to these studies that focus on identifying general network features, in this study we focus on network paths by designing input representations that enable ML algorithms to classify them or predict values that are associated to them (i.e., cost, time). 

\subsection{Machine Learning and Urban Networks}
Previous works in the area of urban network networks are focused on spatio-temporal modeling (\cite{DBLP:conf/ijcai/YuYZ18,Wang2018GraphBasedDM}): 
\cite{DBLP:conf/ijcai/YuYZ18} proposed a deep learning framework for traffic prediction, integrating graph convolution and gated temporal convolution through convolutional blocks. Their framework captures comprehensive spatio-temporal correlations through modeling multi-scale traffic networks. 
\cite{Wang2018GraphBasedDM} designed a generic DNN framework, used for traffic analysis and forecasting, with focus on data that is sparse in both space and time. They proposed a feed forward multilayer perceptron with each node and each edge associated with LSTM cascades instead of weights and activation functions. The estimation of fares is indeed of great interest for efficient trip planning \cite{noulas2015mining}. The problem of taxi fare prediction using Deep Learning approaches is studied in~\cite{rossi2019modelling}.
Compared to this body of work, our focus is on the representation of the paths for training a wide range of ML models, instead of on the design of s specific framework. Moreover, we compress the information before the training of the models. This allows researchers to deal with larger networks that could not otherwise analyzed. 

%

\section{Conclusions}

In this paper we proposed six novel path input representations of urban network paths that can be used as inputs of ML algorithms for classification and regression problems. 
Since urban networks are usually very large in size, we designed input representations that could compress the size of the paths while aiming at preserving as much information as possible. 

We examined the effectiveness of the designed input representations considering a real world case-study. More specifically, we used these representations as inputs for random
decision forest and deep neural network models to predict tips for taxi journeys in New York City using a publicly available dataset as groundtruth. Our results demonstrated that the best performance (i.e., RMSE of 1.42\$) are obtained exploiting the full path enriched with the origin and destination points of the trips and their neighborhoods (\textit{3-steps k-neighbors} representation).
Our analysis also shows that the proposed input representations are effective for classifying urban networks paths. We observe that the \textit{3-steps k-neighbors} is the best representation for predicting tips of taxi rides.
However, we are also aware that, depending on the path characteristics and the size of the network, the best representations for other prediction tasks may be different. 

Finally, is worth noting that the innovation of this work is not only in the application itself, but it is mainly methodological. The proposed input representations can be applied not only to urban networks, but to a variety of regression and classification problems based on paths in large-scale networks.

\section*{Comments}
Accepted for Publication in Environment and Planning B: Urban Analytics and City Science. To Appear.

\bibliographystyle{plain}

%
%
%

\end{document}